\documentclass[a4paper, 11pt]{article}
\usepackage{epsfig, amssymb, amsbsy, amsmath, amsthm, mathrsfs}
\usepackage{psfrag}
\newtheorem{Thm}{Theorem}
\newtheorem{Def}{Definition}

\DeclareMathOperator*{\argmin}{argmin}

\title{\Large A Mathematical Unification of Geometric Crossovers Defined on Phenotype Space}
\author{{\small Yourim Yoon$^\dag$,~~ Yong-Hyuk Kim$^\ddag$,~~ Alberto
Moraglio$^*$, and~ Byung-Ro Moon$^\dag$}\\
{\small $\dag$ School of Computer Science \& Engineering, Seoul National
University}\\
{\small Sillim-dong, Gwanak-gu, Seoul, 151-744, Korea}\\
{\small Email: {\tt \{yryoon, moon\}@soar.snu.ac.kr}}\\
{\small $\ddag$ Department of Computer Science \& Engineering, Kwangwoon
University}\\
{\small 447-1, Wolgye-dong, Nowon-gu, Seoul, 139-701, Korea}\\
{\small Homepage: {\tt http://soar.snu.ac.kr/\~{ }yhdfly}}\\
{\small Email: {\tt yhdfly@kw.ac.kr}}\\
{\small $*$ Department of Computer Engineering, University of Coimbra}\\
{\small Polo II - Pinhal de Marrocos, Coimbra, 3030-290, Portugal}\\
{\small Email: {\tt moraglio@dei.uc.pt}}
}

\date{September 11, 2008}

\begin{document}


\maketitle

\begin{abstract}
Geometric crossover is a representation-independent definition of
crossover based on the distance of the search space interpreted as a
metric space. It generalizes the traditional crossover for binary
strings and other important recombination operators for the most
frequently used representations.
Using a distance tailored to the problem at hand, the abstract
definition of crossover can be used to design new problem specific
crossovers that embed problem knowledge in the search.
This paper is motivated by the fact that genotype-phenotype mapping
can be theoretically interpreted using the concept of
{\em quotient space} in mathematics.
In this paper, we study a metric transformation, the quotient
metric space, that gives rise to the notion of {\em quotient geometric
crossover}. This turns out to be a very versatile notion. 
We give many example applications of the quotient geometric crossover.
\end{abstract}

\section{Introduction}

Geometric crossover and geometric mutation are
representation-independent search operators that generalize many
pre-existing search operators for the major representations used in
evolutionary algorithms, such as binary strings \cite{MorP04-p},
real vectors \cite{YooKAM07}, permutations \cite{MorP05-p}, permutations with
repetitions \cite{KYMM2006}, 
syntactic trees \cite{MorP05-2-p}, and sequences
\cite{moraglio2006a-p}. They are defined in geometric terms using
the notions of line segment and ball. These notions and the
corresponding genetic operators are well-defined once a notion of
distance in the search space is defined. Defining search operators
as functions of the search space is opposite to the standard way
\cite{Jones95} in which the search space is seen as a function of
the search operators employed. This viewpoint greatly simplifies the
relationship between search operators and fitness landscape and has
allowed us to give simple {\em rules-of-thumb} to build crossover
operators that are likely to perform well.

Theoretical results of metric spaces can naturally lead to
interesting results for geometric crossover. In particular, in
previous work \cite{MorP2006} we have shown that
the notion of \emph{metric transformation} has great potential for
geometric crossover. A metric transformation is an operator that
constructs new metric spaces from pre-existing metric spaces: it
takes one or more metric spaces as input and outputs a new metric
space. The notion of metric transformation becomes extremely
interesting when considered together with distances firmly rooted in
the syntactic structure of the underlying solution representation
(e.g., edit distance). In these cases it gives rise to a simple and
\emph{natural interpretation in terms of syntactic transformations}.

In previous work \cite{MorP2006} we have extended
the geometric framework introducing the notion of product crossover
associated with the Cartesian product of metric spaces. This is a
very important tool that allows one to build new geometric
crossovers customized to problems with mixed representations by
combining pre-existing geometric crossovers in a straightforward
way. Using the product geometric crossover, we have also shown that
traditional crossovers for symbolic vectors and blend crossovers for
integer and real vectors are geometric crossover.

In this paper we extend the geometric framework introducing the
important notion of \emph{quotient geometric crossover}. The metric
transformation associated with it is the {\em quotient metric space}.
Quotient space can be regarded as a mathematical definition of
phenotype space in the evolutionary computation theory.
The quotient geometric space has the effect of reducing the search space
actually searched by geometric crossover,
and it introduces problem knowledges in the
search by using a distance better tailored to the specific solution
interpretation.
Quotient geometric crossover is directly applied to the genotype space,
but it has the same effect as the crossover performed on phenotype space.

The paper is organized as follows.
In Section~\ref{sec:geometric_framework}, we present the
geometric framework including the notion of geometricity-preserving
transformation. In Section \ref{sec:qgx}, we introduce the notion of quotient
geometric crossover.
In Section \ref{sec:applications}, we study several useful applications related to
quotient geometric crossover.
In Section \ref{sec:grouping} and \ref{sec:graph},
we show how groupings \cite{KYMM2006} and graphs
can be recast and understood more simply in terms of
quotient geometric crossover. Here, quotient geometric crossover is
used to filter out inherent redundancy in the solution
representation.
In Section \ref{sec:sequence}, 
we show how homologous
crossover for variable-length sequences \cite{moraglio2006a-p}
can be understood as a quotient geometric crossover. 
In Section \ref{sec:TSP}, we discuss the usage of the quotient geometric crossover 
for the traveling salesman problem.
In Section \ref{sec:function}, we consider functional representation
and show how the concept of quotient geometric crossover is
connected to the search of the functions. Genetic programming, finite states
machines, and neural networks are shown as examples.
We explain that quotient geometric crossover can be used
to understand how crossover and neutral code interact in Section
\ref{sec:neutrality}.
In Section~\ref{sec:conclusions}, we give conclusions.

\section{Geometric Framework}
\label{sec:geometric_framework}

\subsection{Geometric Preliminaries}
In the following we give
necessary preliminary geometric definitions and extend those
introduced in \cite{MorP04-p,MorP05-2-p}. The following
definitions are taken from \cite{Deza}.

The terms \emph{distance} and \emph{metric} denote any real-valued
function that conforms to the axioms of identity, symmetry, and
triangular inequality.
In a metric space $(S,d)$ a \emph{line segment} (or closed interval)
is the set of the form $[x;y]_d = \{z \in S ~|~ d(x,z) + d(z,y) =
d(x,y)\}$ where $x,y \in S$ are called extremes of the segment.
Metric segment generalizes the familiar notions of segment in the
Euclidean space to any metric space through distance redefinition.
Notice that a metric segment does not coincide to a shortest path
connecting its extremes (\emph{geodesic}) as in an Euclidean space.
In general, there may be more than one geodesic connecting two
extremes; the metric segment is the union of all geodesics.

We assign a structure to the solution set $S$ by endowing it with a
notion of distance $d$. $M=(S, d)$ is therefore a solution
\emph{space} and $(M, f)$ is the corresponding fitness landscape,
where $f$ is the fitness function over $S$.

\subsection{Definition of Geometric Crossover}

The following definitions are \emph{representation-independent}
therefore applicable to any representation.

\begin{Def}[Image set]
The \emph{image set} $Im[OP]$ of a genetic operator $OP$ is the set
of all possible offspring produced by $OP$.
\end{Def}

\begin{Def}[Geometric crossover]
A binary operator $GX$ is a {\em geometric crossover} under the metric $d$ if
all offspring are in the segment between its parents $x$ and $y$, i.e.,
$$Im[GX(x,y)] \subseteq [x;y]_d.$$
\end{Def}

A number of general properties for geometric crossover and geometric mutation
have been derived in \cite{MorP04-p}.
Traditional crossover is geometric under Hamming distance.
Among crossovers for permutations,
PMX, a well-known crossover for
permutations, is geometric under swap distance. Also, we found that
cycle crossover, another traditional crossover for permutations, is
geometric under swap distance and under Hamming distance.

\subsection{Formal Evolutionary Algorithm and Problem Knowledge}

Geometric operators are defined as functions of the distance
associated with the search space. However, the search space does not
come with the problem itself. The problem consists only of a fitness
function to optimize, that defines what a solution is and how to
evaluate it, but it does not give any structure on the solution set.
The act of putting a structure over the solution set is part of the
search algorithm design and it is a designer's choice.

A fitness landscape is the fitness function plus a structure over
the solution space. So, for each problem, there is one fitness
function but as many fitness landscapes as the number of possible
different structures over the solution set. In principle, the
designer could choose the structure to assign to the solution set
completely independently from the problem at hand. However, because
the search operators are defined over such a structure, doing so
would make them  decoupled from the problem at hand, hence turning
the search into something very close to random search.

In order to avoid this one can exploit problem knowledge in the
search. This can be achieved by carefully designing the connectivity
structure of the fitness landscape. For example, one can study the
objective function of the problem and select a neighborhood
structure that couples the distance between solutions and their
fitness values. Once this is done problem knowledge can be exploited
by search operators to perform better than random search, even if
the search operators are problem-independent (as is the case of
geometric crossover and geometric mutation). Indeed, the fitness landscape is
a knowledge interface between the problem at hand and a formal,
problem-independent search algorithm.

Under which conditions is a landscape  well-searchable by geometric
operators? As a rule of thumb, geometric mutation and geometric
crossover work well on landscapes where the closer pairs of
solutions are, the more correlated their fitness values are. Of course this
is no surprise: the importance of landscape smoothness has been
advocated in many different context and has been confirmed in
uncountable empirical studies with many neighborhood search
meta-heuristics \cite{pardalos2002}. We operate according to the
following rule-of-thumbs:

\noindent \emph{Rule-of-thumb 1}: if we have a good distance for the
problem at hand, then we have a good geometric mutation and a good
geometric crossover.

\noindent \emph{Rule-of-thumb 2}: a good distance for the problem at
hand is a distance that makes the landscape ``smooth.''

\subsection{Geometricity-Preserving Transformation}

In previous work we have proven that a number of important
pre-existing recombination operators for the most frequently used
representations are geometric crossovers. We have also applied the
abstract definition of geometric crossover to distances firmly
rooted in a specific solution representation and designed brand-new
crossovers. An appealing way to build new geometric crossovers is
starting from recombination operators that are known to be geometric
and deriving new geometric crossovers by \emph{geometricity-preserving
transformations/combinations} that when applied to geometric
crossovers, return geometric crossovers.

The definition of geometric crossover is based on the notion of
metric. Therefore, a natural starting point to seek
geometricity-preserving transformations is to consider
transformations of the underlying metrics that are known to return
metric spaces and study how the geometric crossover associated with
the transformed metric space relates with the geometric crossover
associated with the original metric space.

There are a number of metric space transformations \cite{Deza, Sutherland}
that are potentially of interest for geometric
crossover: sub-metric space, product space, quotient metric space,
gluing metric space, combinatorial transformation, non-negative
combinations of metric spaces, Hausdorff transformation, and concave
transformation.

\begin{figure}
\centering
\centerline{\psfig{figure=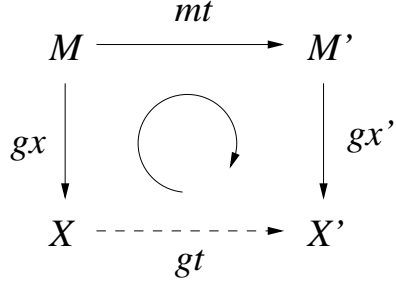,width=2.0in,angle=0}}
\caption[]{Commutative diagram linking metric and crossover
transformations.} \label{fig:comm_diag}
\end{figure}

Let us consider the geometric crossover $X$ associated with the
original metric space $M$, and the geometric crossover $X'$
associated with the transformed metric space $M'=mt(M)$ where $mt$ is
the metric transformation. The functional relationship among metric
spaces and geometric crossovers can be nicely expressed through a
commutative diagram (Figure~\ref{fig:comm_diag}). $gx$ means application of the
formal
definition of geometric crossover and $gt$ means \emph{induced
geometricity-preserving} crossover transformation associated with the
metric transformation $mt$. This diagram becomes remarkably
interesting when the metric transformation $mt$ is associated with an
induced geometricity-preserving crossover transformation $gt$ that
has a simple interpretation in terms of syntactic manipulation. This
indeed allows one to get new geometric crossovers starting from
recombination operators that are known to be geometric by simple
\emph{geometricity-preserving syntax manipulation}.

We study those metric-preserving transformations which induced
geometricity-preserving transformations have a simple and natural
interpretation on the solution representation.

\section{Quotient Geometric Crossover}
\label{sec:qgx}

\subsection{Quotient Metric Space}

Let $(S, d)$ be a metric space and $\sim$ be an equivalence relation on $S$.
Consider the {\em quotient space} $S/\sim$. Now we will give a metric on
$S/\sim$ induced by the original metric $d$ on $S$.
\begin{Def}[Induced distance measure]~\\
For $\bar{x}, \bar{y} \in S/\sim$,
$$d_\sim (\bar{x}, \bar{y}) := \inf_{x \in \bar{x}, y \in \bar{y}} d(x,y).$$
\end{Def}
Then, the following theorem holds \cite{Burago}.
\begin{Thm}
If the equivalence relation arises from an {\em isometry
subgroup}\footnote{\scriptsize For
details, see \cite{Burago}.}, $d_\sim$ is a metric on $S/\sim$.
\end{Thm}
This metric space $(S/\sim, d_\sim)$ is called {\em quotient metric space}.
Later we will directly prove
that $d_\sim$ becomes a metric instead of showing that its related equivalence
relation
$\sim$ comes from an isometry subgroup.

In a metric space $(S, d)$ a \emph{quotient line segment}
is the set of the form $[x;y]_{d_\sim} = \{z \in S ~|~ d_\sim(\bar{x},\bar{z})
+ d_\sim(\bar{z},\bar{y}) = d_\sim(\bar{x},\bar{y}), \bar{z} \in S/\sim\}$
where $\bar{x},\bar{y} \in S/\sim$.
Now we can define quotient geometric crossover.
\begin{Def}[Quotient geometric crossover]~\\
A binary operator $GX_q$ is a {\em quotient geometric crossover}
under the metric $d$ and the equivalence relation $\sim$ if
all offspring are in the quotient line segment between its parents $x$ and $y$,
i.e., $GX_q(x,y) \subseteq [x;y]_{d_\sim}$.
\end{Def}

\subsection{Genotype-Phenotype Mapping}

The notion of quotient geometric crossover is important because it lies
at the heart of the relation between geometric crossover and
genotype-phenotype mapping as we illustrate in the following.

Genotype means solution representation: some structure that can be
stored in a computer and manipulated. Phenotype means solution
itself without any reference to how it is represented. Sometimes it
is possible to have a one-to-one mapping between genotypes and phenotypes,
so the distinction between genotype and phenotype becomes purely
formal. However in many interesting cases phenotypes cannot be
represented uniquely by genotypes. So the same phenotype is
represented by more than one genotypes. In this case we say that we
have a \emph{redundant representation}. For example, to represent a
graph we need to label its nodes and then we can represent it using
its adjacency matrix. This representation is redundant: the same
graph can be represented with more than one adjacency matrix by
relabeling its nodes.

There are quite a few problems in that it is hard to represent one phenotype
by just one genotype using traditional representations.
Roughly speaking, redundant representation leads to severe loss of search power
in genetic algorithms, in particular, with respect to traditional crossovers
\cite{CM03}.
To alleviate the problems caused by redundant representation,
a number of methods such as adaptive crossover have been proposed \cite{DH98,
Las91, Muhl92, HNG02}.
Among them, a technique called {\em normalization}\footnote{\scriptsize
The term of {\em normalization} is firstly appeared in \cite{KM00-2}.
However, it is based on the adaptive crossovers proposed in \cite{Las91,
Muhl92}.}
is representative. It transforms the genotype of a parent to another genotype
to be consistent
with the other parent so that the genotype contexts of the parents are as
similar as possible in crossover.
There have been a number of successful studies using normalization.
An extensive survey about normalization is appeared in \cite{CM03}.

While previous crossovers are usually defined on the subset of genotypes for normalization,
quotient geometric crossover is formally defined on the whole set of genotypes
but actually has the normalization effect.

Although many of studies about normalization did not use the concept of
distance,
once a distance $d_G$ on the genotypes $G$ is defined,
we can formally redefine the normalization $p_2'$ of the second parent $p_2$
to the first $p_1$ as follows:
$$p_2' := \argmin_{s \in w(p_2)} d_G(p_1, s),$$
where $w(s)$ is the set of all the genotypes with the same phenotype as the
genotype $s$.
The use of distance to define normalization is important because it
generalizes and makes rigorous the notion of normalization for
\emph{any solution representation}.

Now we formally present the general relation between geometric crossover and
genotype-phenotype mapping. The concept of normalization defined by distance is
closely related to the quotient geometric crossover.
Let us consider genotype-phenotype mappings $q: G \rightarrow P$
that are not injective (redundant representation). The mapping $q$
induces a natural equivalence relation $\sim$ on the set of
genotypes: {\em genotypes with the same phenotype belong to the same
class}. Given a distance $d_G$ on genotypes $G$, the quotient with
the relation $\sim$ produces a distance $d_P$ on the phenotypes $P$:
$P=G/\sim$ and $d_P(\bar{x}, \bar{y}) = \inf_{x \in \bar{x}, y \in
\bar{y}} d_G(x,y)$.

By applying the formal definition of geometric crossover to the
metric spaces $(G,d_G)$ and $(P,d_P)$, we obtain the geometric
crossovers $X_G$ and $X_P$, respectively. $X_G$ searches the space of
genotypes and $X_P$ searches the space of phenotypes. Searching the
space of phenotypes has a number of advantages: (i) it is smaller
than the space of genotypes, hence quicker to search (ii) the
phenotypic distance is better tailored to the underlying problem,
hence the corresponding geometric crossover works better (iii) the
space of phenotypes has different geometric characteristics from the
genotype space. This can be used to remove unwanted bias from
geometric crossover.

\begin{figure}
\centering
\centerline{\psfig{figure=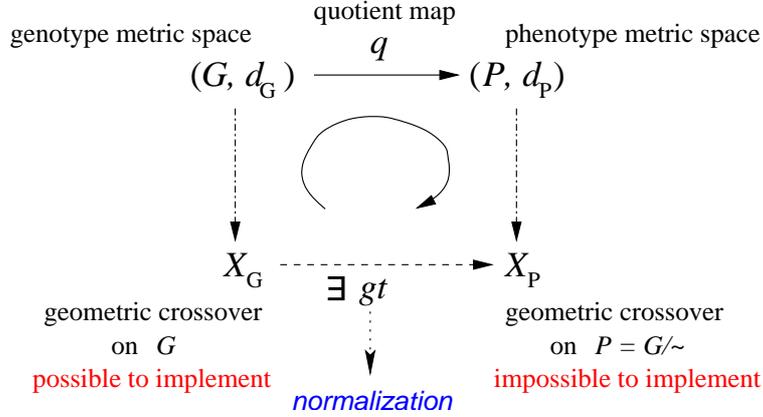,width=4.0in,angle=0}}
\caption[]{Functional relationship among genotype \& phenotype
metric spaces and geometric crossovers on them}
\label{fig:comm_diag2}
\end{figure}

However, the crossover $X_P$ cannot be directly used itself because
it recombines phenotypes that are objects that cannot be directly
represented. The quotient geometric crossover allows us to search
the space of phenotypes with the crossover $X_P$ \emph{indirectly}
by manipulating the genotypes $G$. This is possible because for the
commutative diagram there exists an induced geometricity-preserving
transformation $gt$ of the genotypic crossover $X_G$ that allows us
to use the genotypic representation to implement a geometric
crossover in the space of phenotypes $gt(X_G)$ without making
explicit use of phenotypes (see Figure~\ref{fig:comm_diag2}).
The type of the transformation $gt$
depends on the type of the equivalence relation $\sim$ used in the
quotient of the underlying metric space that in turns depends on
the underlying syntax of the solution representation. It may happen
that the induced geometricity-preserving transformation may be
difficult to implement and/or computationally intractable. In these
cases, it may not be feasible using an exact equivalent of the
phenotypic geometric crossover, but an approximation may be
preferable and still retaining most of the advantages of the exact
equivalent.

In the following section we consider a number of equivalence
classes for the quotient operation and its related induced genotypic
crossover transformation.

\section{Applications}
\label{sec:applications}

Quotient geometric crossover has various applications.
Although some of these methods have already been used independently,
we unify the methods, which look quite different,
under the concept of quotient geometric crossover.

\subsection{Groupings}
\label{sec:grouping}

{\em Grouping} problems \cite{Falk98} are commonly
concerned with partitioning given item set into mutually disjoint
subsets. Examples belonging to this class of problems are multiway
graph partitioning, graph coloring, bin packing, and so on. Grouping
representation is also used to solve the joint replenishment
problem, which is a well-known problem appeared in the field of
industrial engineering \cite{Olsen05}. In this class of problems,
the normalization decreased the problem difficulty and led to
notable improvement in performance.

Most normalization studies for grouping problems were focused on the $k$-way
partitioning problem. In the problem, the $k$-ary representation, in which $k$
subsets are represented by the integers from $0$ to $k - 1$, has been generally
used. In this case, a phenotype (a $k$-way partition) is represented by $k!$
different genotypes. In the problem, a normalization method was used in
\cite{KM00-2}.
Other studies for the $k$-way partitioning problem used the same technique
\cite{CM03, JPKim01}.
In sense that normalization pursues the minimization of genotype inconsistency
among chromosomes, in previous work \cite{KimM05-p}, we proposed
an optimal, efficient normalization method for grouping problems and
a distance measure, the {\em labeling-independent distance}, that eliminates
this dependency completely.

Let $\mathfrak{a},\mathfrak{b} \in U = \{1, 2, \ldots, k\}^n$ 
be $k$-ary encodings (fixed-length vectors on a $k$-ary alphabet)
and $H$ be the Hamming distance in $U$.
We define $\mathfrak{a}$ and $\mathfrak{b}$
to be in relation $\sim$ if
there exist $\sigma$ and $\sigma'$ in $\Sigma_k$ such that
$\mathfrak{a}_\sigma = \mathfrak{b}_{\sigma'}$
where $\Sigma_k$ is the set of all permutations of length $k$ and
$\mathfrak{a}_\sigma$ is a permuted encoding of $\mathfrak{a}$ by a
permutation $\sigma$, i.e., the $i^{th}$ element $a_i$ of
$\mathfrak{a}$ is transformed into $\sigma(a_i)$.
Then,
the relation $\sim$ is an equivalence relation (see \cite{KimM05-p}).

We define the {\em labeling-independent distance}
$LI$ on $U/\sim$ as follows:
\begin{displaymath}
LI(\bar{\mathfrak{a}},\bar{\mathfrak{b}}) := \min_{\sigma, \sigma' \in \Sigma_k}
H(\mathfrak{a}_\sigma,\mathfrak{b}_{\sigma'})
\end{displaymath}
$(U/\sim,LI)$ is a metric space, i.e., the labeling-independent
distance $LI$ is a metric on $U/\sim$ (see \cite{KimM05-p}).

We designed a new crossover based on the
labeling-independent metric in previous work \cite{KYMM2006}.
\begin{Def}[Labeling-independent crossover]
Normalize the second parent to the first under the Hamming distance $H$.
Do the normal crossover using the first parent and the normalized
second parent.
\end{Def}

In fact, this crossover is the quotient geometric crossover
since its offspring are exactly on quotient line segment.
We proved it in \cite{KYMM2006} though we did not 
represented with the notion of quotient geometric crossover.
In sum, we have:

\begin{center}
\begin{tabular}{|c|c|} \hline
original     & labeled partitions \\
metric space & under Hamming distance \\ \hline
quotient     & unlabeled partitions \\
metric space & under labeling-independent distance \\ \hline \hline
original        & traditional crossover\\
geometric crossover & \\ \hline
quotient        & label normalization \\
geometric crossover & before traditional crossover \\ \hline
\end{tabular}
\end{center}

\begin{itemize}
    \item \emph{Genotypes} $G$: labeled partitions represented as
    vectors of symbols
    \item \emph{Phenotypes} $P$: unlabeled partitions
    \item \emph{Equivalence relation} $\sim$: labeled partitions
    with the same partition structure
    \item \emph{Distance on genotypes} $d_G$: Hamming distance
    \item \emph{Distance on phenotypes} $d_P$: labeling-independent
distance
    \item \emph{Crossover on genotypes} $X_G$: traditional crossover
    for vectors
    \item \emph{Crossover of phenotypes} $X_P$: label normalization
    before traditional crossover
    \item \emph{Induced crossover transformation} $gt$: label
    normalization
\end{itemize}

The benefit of understanding normalization for grouping problems in
terms of quotient geometric crossover is the possibility of
understanding the benefit of normalization in terms of landscape
analysis. We have done this in previous work \cite{KYMM2006}.

\subsection{Graphs}
\label{sec:graph}

In this subsection, we consider any problem naturally defined over a graph in
which the
fitness of the solution does not depend on the labels on the nodes but
only on the structural relationship, i.e., edge between nodes.

Formally, let $A \in \mathfrak{M}_n$ be the adjacency matrix of a labeled graph
using labels of $n$ nodes and let $P$ be an $n \times n$
permutation matrix\footnote{ \scriptsize
Permutation matrix is a $(0,1)$-matrix with exactly one $1$ in every row and
column.}.
Then the matrix $PA$ means the labeled graph obtained by relabeling $A$
according to
the permutation represented by $P$.
The fitness $f: \mathfrak{M}_n \rightarrow \mathbb{R}$
satisfies that for every $A \in \mathfrak{M}_n$ and every permutation matrix
$P$,
$f(A)=f(PA)$.

Let $(\mathfrak{M}_n,H)$ be a metric space on the labeled graphs under the
Hamming distance $H$.
Notice that this metric is labeling-dependent.
In particular, $H(A,PA)$ may not be zero
although $A$ and $PA$ represent the same structure.
If $A$ is equal to $PA'$ for some permutation matrix $P$,
we define $A$ and $A'$ to be {\em in relation $\sim$}, i.e., $A \sim A'$.
Then, the relation $\sim$ is an equivalence relation.

An {\em unlabeled graph} $\mathfrak{g}$ is the equivalence class of all its
labeled graphs,
i.e.,\\ $\mathfrak{g}(A) = \{PA~|~P \textrm{ is a permutation matrix} \}$.
{\em unlabeled-graph space} $\mathfrak{M}_n/\sim$ is the set of all equivalence
classes
partitioning the set $\mathfrak{M}_n$.

We define {\em induced distance measure} $LI$ on $M_n/\sim$ as follows:
for each $\mathfrak{g}, \mathfrak{g}' \in \mathfrak{M}_n/\sim$,
$$LI(\mathfrak{g},\mathfrak{g}'):=\min_{A \in \mathfrak{g}, A' \in
\mathfrak{g}'} H(A,A').$$
Then, $(\mathfrak{M}_n/\sim,LI)$ is a metric space, i.e., $LI$ is a metric on
$\mathfrak{M}_n/\sim$.
It shows that the metric space $(\mathfrak{M}_n,H)$ induces a quotient metric
space
$(\mathfrak{M}_n/\sim,LI)$.

\begin{Def}[Labeling-independent crossover]
Do the graph matching of the second parent $p_2$ to the first $p_1$ under the
Hamming distance $H$,
i.e., $$p_2' := \argmin_{A \in \mathfrak{g}(p_2)} H(p_1, A).$$
Do the normal crossover using the first parent $p_1$ and the graph-matched
second parent $p_2'$.
\end{Def}

\noindent
The following theorem shows that the labeled-graph geometric crossover
for $(\mathfrak{M}_n,H)$ induces the unlabeled-graph geometric crossover
for $(\mathfrak{M}_n/\sim,LI)$.

\begin{Thm}\label{thm:gx}
The labeling-independent crossover is geometric under the metric $LI$.
\end{Thm}

The labeling-independent crossover is defined over unlabeled graphs
$\mathfrak{M}_n/\sim$.
This space is much
smaller than labeled graphs $\mathfrak{M}_n$. More precisely,
$|\mathfrak{M}_n/\sim|=|\mathfrak{M}_n|/n!$. This means that the more the
labels are, the
smaller the unlabeled-graph space is compared with the labeled-graph space.
Smaller space means better performance given the same amount of
evaluations.

The previous theorem tells how to guide the implementation
using graph matching for specific geometric crossovers.
To implement the geometric crossover over unlabeled graphs, we need to use
labeled graphs.
The labeling results are necessary to represent and handle the solution,
even if in fact it is only an auxiliary function and can be considered
not being part of the problem to solve. Graph matching before crossover
allows to implement the geometric crossover on the unlabeled-graph
space using the corresponding geometric crossover over the auxiliary
space of the labeled graph after graph matching.
In sum, we have:

\begin{center}
\begin{tabular}{|c|c|} \hline
original     & labeled graphs \\
metric space & under Hamming distance \\ \hline
quotient     & unlabeled graphs \\
metric space & under labeling-independent distance \\ \hline \hline
original        & traditional crossover\\
geometric crossover & \\ \hline
quotient        & graph matching \\
geometric crossover & before traditional crossover \\ \hline
\end{tabular}
\end{center}

\begin{itemize}
    \item \emph{Genotypes} $G$: labeled graphs with the same number of
nodes represented as
    adjacency matrices of the same size
    \item \emph{Phenotypes} $P$: unlabeled graphs
    \item \emph{Equivalence relation} $\sim$: adjacency matrices
    with the same underlying unlabeled graph
    \item \emph{Distance on genotypes} $d_G$: Hamming distance
    between adjacency matrices
    \item \emph{Distance on phenotypes} $d_P$: labeling-independent
    distance between unlabeled graphs. This equals the \emph{edge edit
    distance}.
    \item \emph{Crossover on genotypes} $X_G$: traditional crossover
    on adjacency matrices seen as vectors
    \item \emph{Crossover of phenotypes} $X_P$: graph
    matching before traditional crossover on adjacency matrices
    \item \emph{Induced crossover transformation} $gt$: graph
    matching
\end{itemize}

The benefit of applying the quotient geometric crossover on graphs
is the design of a crossover better tailored to graphs. The notion
of graph matching before crossover arises directly from the
definition of quotient geometric crossover. Graphs are very important because
they are ubiquitous. In future work we will test this crossover on some
applications. Graphs and groupings can be seen as particular cases of
labeled structures in which the fitness of a solution depends only
on the structure and not on the specific labeling. In future work we
will study the class of labeled structures in combination with
quotient geometric crossover.

\subsection{Sequences}
\label{sec:sequence}

In this subsection we recast alignment before recombination in
variable-length sequences as a consequence of quotient geometric
crossover. In previous work \cite{moraglio2006a-p} we have applied
geometric crossover to variable-length sequences. The distance for
variable-length sequences we used there is the {\em edit distance}
$LD$\footnote{\scriptsize The notation $LD$ comes from
{\em Levenshtein distance} that is another name of edit distance.}:
the minimum number of insertion, deletion, and replacement of single
character to transform one sequence into the other. The geometric
crossover associated with this distance is the {\em homologous
geometric crossover}: two sequences are aligned optimally before
recombination. Alignment here means allowing parent sequences to be
{\em stretched} to match better with each other. Formally stretching
sequences means interleaving `-' anywhere and in any number in the
sequences to create two stretched sequences of the same length that
have minimal Hamming distance. For example, if we want to recombine
{\tt agcacaca} and {\tt acacacta}, we need to align them optimally
first: {\tt agcacac-a} and {\tt a-cacacta}. Notice that the Hamming
distance between the aligned sequences is less than the Hamming distance
between the non-aligned sequences.

After the optimal alignment, one does the normal crossover
and produces a new {\em stretched} sequence. The
offspring is obtained by removing `-', so by unstretching the sequence.
How does quotient geometric crossover fit in here?
We can define a relation $\sim$ on stretched sequences: {\em each
stretched sequence belongs to the class of its unstretched version}.
Then, we can easily check that the relation $\sim$ is an equivalence relation.
Let $\langle s \rangle$ be the set of all stretched sequences of sequence $s$.
We define the induced distance measure $d_\sim$.
Let $s_1, s_2$ be variable-length sequences.
If $H$ is the Hamming distance for stretched sequences,
\begin{displaymath}
d_{\sim}(s_1, s_2) := \min_{s'_1 \in \langle s_1 \rangle, s'_2 \in \langle s_2
\rangle} H(s'_1, s'_2).
\end{displaymath}
Then, by the definition of edit distance, $d_\sim$ is equal to $LD$.
Hence $d_\sim$ is a metric on variable-length sequences.

\begin{Thm}
Homologous crossover is geometric under the edit distance
\cite{moraglio2006a-p}.
\end{Thm}


\noindent In summary, we have the following.
\begin{center}
\begin{tabular}{|c|c|} \hline
original     & stretched sequences \\
metric space & under Hamming distance$^\dag$ \\ \hline
quotient     & sequences \\
metric space & under edit distance \\ \hline \hline
original        & traditional crossover\\
geometric crossover & \\ \hline
quotient        & homologous crossover \\
geometric crossover & \\ \hline
\end{tabular}
\end{center}
{\scriptsize $\dag$ If sequences have different length,
their Hamming distance is applied after
aligning the sequences leftmost, and the tail of the longer sequence is
considered different from the missing tail of the shorter sequence.}\\
This idea can be extended to any stretchable structure, e.g., {\em stretchable
graphs}.

\begin{itemize}
    \item \emph{Genotypes} $G$: variable-length stretched sequences
    \item \emph{Phenotypes} $P$: variable-length (unstretched)
    sequences
    \item \emph{Equivalence relation} $\sim$: stretched sequences with the same
    unstretched sequence
    \item \emph{Distance on genotypes} $d_G$: If the two
    stretched sequences have different length, add as many `-' as necessary at
the right end of the shorter
    sequence to make it become equal in length to the longer
    sequence. Their genotypic distance is then their Hamming distance.
    \item \emph{Distance on phenotypes} $d_P$: edit distance between
    sequences
    \item \emph{Crossover on genotypes} $X_G$: traditional crossover
    on stretched sequences. If the two
    stretched sequences have different length, add as many `-' as necessary at
the right end of the shorter
    sequence to make it become equal in length to the longer
    sequence.
    \item \emph{Crossover of phenotypes} $X_P$: homologous crossover
    for sequences
    \item \emph{Induced crossover transformation} $gt$: optimal alignment
\end{itemize}

Phenotypes are variable-length sequences that are directly
representable. So in this case the quotient geometric crossover is
not used to search a non-directly representable space (phenotypes)
through an auxiliary directly representable space (genotypes). The
benefit of applying the quotient geometric crossover on
variable-length sequences is that the homologous crossover over
sequences $X_P$ is naturally understood as a transformation $gt$ of
the geometric crossover $X_G$ over stretched sequences $G$ rather
than a crossover acting directly on sequences $P$. This is because
the notion of optimal alignment is inherently defined on stretched
sequences and not on simple sequences. In previous work \cite{moraglio2006a-p}
we have tested the homologous crossover on the protein motif
discovery problem. In future work we want to study how the optimal
alignment transformation affects the fitness landscape associated
with geometric crossover with and without alignment.

\subsection{Traveling Salesman Problem}
\label{sec:TSP}

In previous work \cite{MorP05-2-p} we have applied geometric
crossover to traveling salesman problem (TSP).
Solutions are tours of cities, or circular
permutations. A good neighborhood structure for TSP is the one
based on the 2-opt move. This move simply reverses the order of the
cities of a contiguous subtour. This move induces a graphic distance
between tours: the minimum number of reversals to transform one tour
into the other. The geometric crossover associated with this
distance belongs to the family of sorting crossovers: it picks
offspring on the minimum sorting trajectory between parent circular
permutations sorted by reversals. Tours of cities or circular
permutations cannot be represented directly. They are represented
with simple permutations. Gluing head and tail of the permutation
obtains a circular permutation. However each circular permutation is
represented by more than one simple permutation. How does quotient
geometric crossover fit in here? We can define an equivalence
relation on the simple permutations: {\em each simple permutation
belongs to the class of its associated circular permutation}. So we
have:
\begin{center}
\begin{tabular}{|c|c|} \hline
original     & simple permutations \\
metric space & under reversal distance \\ \hline
quotient     & circular permutations \\
metric space & under reversal distance \\\hline \hline
original        & sorting by reversal crossover \\
geometric crossover & for simple permutations \\\hline
quotient        & sorting by reversal crossover \\
geometric crossover & for circular permutations \\ \hline
\end{tabular}
\end{center}

\begin{itemize}
    \item \emph{Genotypes} $G$: permutations
    \item \emph{Phenotypes} $P$: circular permutations (tours)
    \item \emph{Equivalence relation} $\sim$: permutations
    identifying the same circular permutation
    \item \emph{Distance on genotypes} $d_G$: reversal distance
    between permutations
    \item \emph{Distance on phenotypes} $d_P$: reversal distance
    between circular permutations
    \item \emph{Crossover on genotypes} $X_G$: based on sorting by reversals
    for permutations
    \item \emph{Crossover of phenotypes} $X_P$: based on sorting by reversals
    for circular permutations implemented using simple permutations:
    circular shift to match as much as possible the two simple permutations
    before sorting crossover
    \item \emph{Induced crossover transformation} $gt$: circular
    shift before sorting by reversal crossover
\end{itemize}

This example of quotient geometric crossover illustrates how to
obtain a geometric crossover for a transformed representation
(circular permutation) starting from a geometric crossover for the
original representation (simple permutation). So in this case
quotient geometric crossover is used as a tool to build a new
crossover for a derivative representation from a known geometric
crossover for the original representation. From previous work we
know that the sorting by reversal crossover for simple permutations
is an excellent crossover for TSP. In future work we want to test
the sorting by reversal crossover for circular permutations. Since
they are a direct representation of city tours we expect it to
perform even better.

\subsection{Functions}
\label{sec:function}

Here we consider functional representations: any representation that
encodes a function. Examples of this type of representation are
genetic programming (GP) trees, finite state automata, and neural
networks. We can define an equivalence relation on the solution
space: {\em all solutions representing the same function}. So we
have:
\begin{center}
\begin{tabular}{|c|c|} \hline
original     & original metric \\
metric       & for the specific representation \\ \hline
quotient     & {\em representation-independent metric} \\
metric       & among representable functions \\\hline \hline
original        & geometric crossover \\
geometric crossover & for the specific representation \\\hline
quotient        & geometric crossover \\
geometric crossover & in the function space \\ \hline
\end{tabular}
\end{center}

\subsubsection{Genetic Programming}

We can define an equivalence relation: {\em all symbolic expressions
that represent the same function}. We can also define a less strong
equivalence relation: {\em consider as equivalent those syntactic
trees that differ in the order of the operands in nodes with
commutative operations}. For example, the multiplication operation
`$*$' is commutative and two different trees represent the same
function.
\begin{center}
\begin{tabular}{|c|c|} \hline
original     & structural Hamming distance \\
metric       & between rooted ordered trees \cite{MorP05-cec-H} \\ \hline
quotient     & structural Hamming distance \\
metric       & between rooted unordered trees \\
             & (only for commutative nodes) \\ \hline \hline
original        & homologous crossover \\
geometric crossover & for GP trees \\\hline
                & homologous crossover for GP trees \\
quotient        & with reordering of commutative \\
geometric crossover & subtrees to have minimum \\
                & structural Hamming distance \\ \hline
\end{tabular}
\end{center}
This quotient geometric crossover is based on the less
strong equivalence relation. So it is not {\em fully} semantical. However
already this quotient geometric crossover cannot be implemented
efficiently because the complexity to compute the structural Hamming
distance between rooted unordered trees grows exponentially with the
number of nodes in the trees.

\begin{itemize}
    \item \emph{Genotypes} $G$: parse trees that is a compact (shorter than
extensive
    form), redundant (the same function can be represented by more than one parse
tree)
    and biased (some functions have more associated parse trees than other
functions)
    representation of functions.
    \item \emph{Phenotypes} $P$: computed functions. A generic
    function can be directly represented in an extensive form as a vector in
    which for every combinations of the input values there is a cell
    that contains the output of the function for those values. We
    call this vector the output vector representation of the
    function. Clearly this direct representation in practice is not used
    because it is too long.
    \item \emph{Equivalence relation} $\sim$: parse trees that
    correspond to the same function or equivalently with the same output
vector.
    \item \emph{Distance on genotypes} $d_G$: structural Hamming
    distance between parse trees
    \item \emph{Distance on phenotypes} $d_P$: (weighted) Hamming distance on
output vectors
    \item \emph{Crossover on genotypes} $X_G$: homologous crossover
    for parse trees
    \item \emph{Crossover of phenotypes} $X_P$: traditional
    crossover on the output vectors of the functions
    \item \emph{Induced crossover transformation} $gt$:
    expand/reduce/ change syntactic trees before crossover without changing
    the underlying computed functions such as they have minimal
    structural Hamming distance
\end{itemize}

The benefit of applying quotient geometric crossover to parse trees
is to search the space of the functions represented by the parse
trees rather than the space of parse trees. This is done indirectly
by manipulating parse trees. Even if in principle a function can be
represented directly using its output vector representation, so
making not strictly necessary to recur to an auxiliary genotypic
representation and to the quotient geometric crossover to search
this space, such direct representation is simply too long for any
practical purpose, and a concise genotypic representation is needed.

The implementation of the phenotypic geometric crossover using the
transformation $gt$ on the genotypic crossover $X_G$ presents a
problem: it is simply not possible to compute efficiently the
transformation $gt$ because one needs to compute the smallest
structural Hamming distance between all possible transformations of
the syntactic trees that keep invariant their underlying functions.
We could relax the problem and consider a weaker equivalence
relation in which two parse trees are equivalent if exchanging
subtrees of nodes with commutative operations (syntactic
transformation that keeps the computed function invariant) they
become equal. In this case $d_P$ becomes the distance between rooted
(partially) unordered trees. The computational cost of this distance
grows exponentially with the number of commutative nodes in the
syntactic trees. This could be still hard to compute and so could the
associated geometric crossover $X_P$. However there are quick
approximated algorithms to compute this distance. In future work we
will try this crossover.

\subsubsection{Finite States Machines}

Finite state machines can represent discrete functions or
classifiers: given in input any sequence, they return the class of
this sequence. They are represented as labeled rooted directed
graphs or equivalently with a transition matrix. We can define an
equivalence relation: {\em all the FSMs that represent the same
classifier}. We can also define a less strong equivalence relation:
{\em all the unlabeled FSMs that represent the same classifier}. In
fact, as for graph partitioning the labels are assigned completely
arbitrarily.\\
\begin{center}
\begin{tabular}{|c|c|} \hline
original     & Hamming distance \\
metric       & on transition matrix \\ \hline
quotient     & Hamming distance \\
metric       & on normalized transition matrix \\\hline \hline
original        & traditional crossover \\
geometric crossover & \\\hline
quotient        & normalization before recombination \\
geometric crossover & of the transition matrix \\ \hline
\end{tabular}
\end{center}

\begin{itemize}
    \item \emph{Genotypes} $G$: transition matrices
    \item \emph{Phenotypes} $P$: classification functions
    \item \emph{Equivalence relation} $\sim$: transition matrices
    that give rise to the same classifier (same output vectors defined as for
the parse trees)
    \item \emph{Distance on genotypes} $d_G$: Hamming distance
    between transition matrices
    \item \emph{Distance on phenotypes} $d_P$: minimum Hamming distance
    between unlabeled transition matrices that equals (weighted) Hamming
distance on output vectors
    \item \emph{Crossover on genotypes} $X_G$: traditional crossover
    on transition matrices
    \item \emph{Crossover of phenotypes} $X_P$: crossover of the
    underlying classifiers (traditional crossover on the output vectors)
    \item \emph{Induced crossover transformation} $gt$:
    put both FSMs in a normal form (for example using their lexicographic
    order) before recombination of their transition matrices with
    traditional crossover. This is a quick heuristic that
    approximates the phenotypic crossover.
\end{itemize}

The benefit of the quotient geometric crossover is to be able to
search the space of classifiers using a concise representation.

\subsubsection{Neural Networks}

Neural networks can be represented by real matrices of the
connection weights. We can define an equivalence relation: {\em all
the matrices that give rise to the same input-output mapping}. We
can also define a less strong equivalence relation: {\em all the
matrices that when reordered become the same}.
\begin{center}
\begin{tabular}{|c|c|} \hline
original     & Manhattan distance \\
metric       & on matrices \\ \hline
quotient     & Manhattan distance \\
metric       & between unlabeled matrices \\\hline \hline
original        & traditional box crossover \\
geometric crossover & for real vectors \\\hline
quotient        & normalization before box crossover \\
geometric crossover & for real vectors \\ \hline
\end{tabular}
\end{center}

\begin{itemize}
    \item \emph{Genotypes} $G$: weights matrices
    \item \emph{Phenotypes} $P$: continuous functions
    \item \emph{Equivalence relation} $\sim$: weights matrices are equivalent
when giving rise to the same function
    (the same output vectors defined as for the parse trees, in this case the
vector is infinite-dimensional)
    \item \emph{Distance on genotypes} $d_G$: Manhattan distance
    between weights matrices
    \item \emph{Distance on phenotypes} $d_P$: weight-label-independent
Manhattan distance between weights matrices
    equals properly weighted Manhattan distance on output vectors (distance
must be a finite number)
    \item \emph{Crossover on genotypes} $X_G$: box recombination of
    weights matrices
    \item \emph{Crossover of phenotypes} $X_P$: box recombination
    continuous functions
    \item \emph{Induced crossover transformation} $gt$:
    normalization on weight-label before box recombination of the
    weights matrices
\end{itemize}

The benefit of the quotient geometric crossover is to be able to
search with its geometric crossover the space of continuous function
indirectly using a concise representation (the weights matrices
representation). The geometric crossover defined over continuous
function cannot be implemented directly on the phenotype space
(space of functions) because it would need to recombine
infinite-dimensional vectors.

\subsection{Neutrality}
\label{sec:neutrality}

The role of neutrality is little understood. Notice that {\em
neutrality} is a synonym of redundancy. As a rule of thumb one would
like to filter out redundancy as in normalization for structural
problem to improve performance. However neutrality may have some
beneficial aspect on variable-length representation. In fact it can
be used to have a self-adaptive mutation rate at a phenotypic level:
imaging you have a constant number of mutations at a genotype level.
If the informative part, the one used to get the phenotype, is small
as compared with the non-informative one, mutation at genotype level
have a small chance to affect the phenotype. So the same mutation
rate at genotype level can correspond to a smaller or equivalent
mutation at a phenotype level depending on the amount of neutral
code in the genotype. Since the mutation itself inserts or deletes
neutral code, this combined with selection develops a self-adaptive
mechanism that selects genotypes with the right amount of neutral
code to be more evolvable. Neutrality is widespread in nature, so
studying neutrality is important. Quotient geometric crossover can
be used to understand how crossover and neutrality interact. In fact
the induced geometricity-preserving transformation tells what {\em
trick} to use to remove redundancy for crossover but still keep it
there for mutation to obtain the self-adaptive mutation rate {\em
trick}, so to take advantage of both genotype and phenotype
spaces.

\begin{itemize}
    \item \emph{Genotypes} $G$: sequence with neutral code
    (part of the sequence that if removed would not affect the phenotype)
    \item \emph{Phenotypes} $P$: sequence without neutral code.
    There is a one-to-one mapping between these sequences and the
    phenotypes. So this is a direct representation of the phenotypes, rather
than the phenotype itself.
    \item \emph{Equivalence relation} $\sim$: two sequences with
    neutral code are equivalent if when the neutral code is removed
    they become the same phenotypic sequence.
    \item \emph{Distance on genotypes} $d_G$: edit distance on
    sequences with neutral code
    \item \emph{Distance on phenotypes} $d_P$: edit distance on
    sequences without neutral code
    \item \emph{Crossover on genotypes} $X_G$: homologous crossover
    for sequences
    \item \emph{Crossover of phenotypes} $X_P$: homologous crossover
    for sequences
    \item \emph{Induced crossover transformation} $gt$:
    identity transformation
\end{itemize}

The benefit of the quotient geometric crossover is to show how
crossover and neutral code interact. We have seen that neutrality
may be beneficial in terms of adaptive mutation rate. Since the
induced crossover transformation is the identity transformation,
this means that the same crossover that searches the genotypes space
can be understood as a crossover searching the phenotype space
indirectly using the genotypes. In other words, the neutral code is
completely transparent to the search done by crossover and it does
not affect its search or performance. So, neutral code retains the
advantage of an adaptive mutation rate together with being
transparent to the action of crossover.

\section{Concluding Remarks}
\label{sec:conclusions}

In this paper we have extended the geometric framework introducing
the notion of quotient geometric crossover.
This could be clearly understood
using the concept of geometricity-preserving transformation.
Quotient geometric crossover is a very general and
versatile tool. We have given a number of interesting examples as its
applications. As shown in applications,
quotient geometric crossover is not only theoretically significant
but also has a practical effect of making search more effective
by reducing the search space or removing the inherent bias.
More theoretical analysis will be appeared in the extended full paper,
and more detailed applications for each case are left for future study.

{
\bibliographystyle{plain}
\bibliography{ref}

\begin{thebibliography}{10}

\bibitem{Burago}
D.~Burago, Y.~Burago, S.~Ivanov, and Iu.~D. Burago.
\newblock {\em A Course in Metric Geometry}.
\newblock American Mathematical Society, 2001.

\bibitem{CM03}
S.-S. Choi and B.-R. Moon.
\newblock Normalization in genetic algorithms.
\newblock In {\em Proceedings of the Genetic and Evolutionary Computation
  Conference}, pages 862--873, 2003.

\bibitem{Deza}
M.~M. Deza and M.~Laurent.
\newblock {\em Geometry of Cuts and Metrics}.
\newblock Springer, 1997.

\bibitem{DH98}
R.~Dorne and J.~K. Hao.
\newblock A new genetic local search algorithm for graph coloring.
\newblock In {\em Proceedings of the Fifth Conference on Parallel Problem
  Solving from Nature}, pages 745--754, 1998.

\bibitem{Falk98}
E.~Falkenauer.
\newblock {\em Genetic Algorithms and Grouping Problems}.
\newblock John Wiley \& Sons, 1998.

\bibitem{Jones95}
T.~Jones.
\newblock {\em Evolutionary Algorithms, Fitness Landscapes and Search}.
\newblock PhD thesis, University of New Mexico, 1995.

\bibitem{KM00-2}
S.-J. Kang and B.-R. Moon.
\newblock A hybrid genetic algorithm for multiway graph partitioning.
\newblock In {\em Proceedings of the Genetic and Evolutionary Computation
  Conference}, pages 159--166, 2000.

\bibitem{JPKim01}
J.-P. Kim and B.-R. Moon.
\newblock A hybrid genetic search for multi-way graph partitioning based on
  direct partitioning.
\newblock In {\em Proceedings of the Genetic and Evolutionary Computation
  Conference}, pages 408--415, 2001.

\bibitem{KimM05-p}
Y.-H. Kim and B.-R. Moon.
\newblock New topologies for genetic search space.
\newblock In {\em Proceedings of the Genetic and Evolutionary Computation
  Conference}, pages 1393--1399, 2005.

\bibitem{Las91}
G.~Laszewski.
\newblock Intelligent structural operators for the $k$-way graph partitioning
  problem.
\newblock In {\em Proceedings of the Fourth International Conference on Genetic
  Algorithms}, pages 45--52, 1991.

\bibitem{KYMM2006}
A.~Moraglio, Y.-H. Kim, Y.~Yoon, and B.-R. Moon.
\newblock Geometric crossover for multiway graph partitioning.
\newblock In {\em Evolutionary Computation}, volume~15, pages 445--474, 2007.

\bibitem{MorP04-p}
A.~Moraglio and R.~Poli.
\newblock Topological interpretation of crossover.
\newblock In {\em Proceedings of the Genetic and Evolutionary Computation
  Conference}, pages 1377--1388, 2004.

\bibitem{MorP05-2-p}
A.~Moraglio and R.~Poli.
\newblock Geometric crossover for the permutation representation.
\newblock {\em Technical Report CSM-429, Department of Computer Science,
  University of Essex}, 2005.

\bibitem{MorP05-cec-H}
A.~Moraglio and R.~Poli.
\newblock Geometric landscape of homologous crossover for syntactic trees.
\newblock In {\em Proceedings of the IEEE Congress on Evolutionary
  Computation}, volume~1, pages 427--434, 2005.

\bibitem{MorP05-p}
A.~Moraglio and R.~Poli.
\newblock Topological crossover for the permutation representation.
\newblock In {\em GECCO 2005 Workshop on Theory of Representations}, 2005.

\bibitem{MorP2006}
A.~Moraglio and R.~Poli.
\newblock Product geometric crossover.
\newblock In {\em Proceedings of the Conference on Parallel Problem Solving
  from Nature}, pages 1018--1027, 2006.

\bibitem{moraglio2006a-p}
A.~Moraglio, R.~Poli, and R.~Seehuus.
\newblock Geometric crossover for biological sequences.
\newblock In {\em Proceedings of European Conference on Genetic Programming},
  pages 121--132, 2006.

\bibitem{Muhl92}
H.~M{\"{u}}hlenbein.
\newblock Parallel genetic algorithms in combinatorial optimization.
\newblock In {\em Computer Science and Operations Research: New Developments in
  Their Interfaces}, pages 441--456, 1992.

\bibitem{Olsen05}
A.~L. Olsen.
\newblock An evolutionary algorithm to solve the joint replenishment problem
  using direct grouping.
\newblock {\em Computers and Industrial Engineering}, 48(2):223--235, 2005.

\bibitem{pardalos2002}
P.~M. Pardalos and M.~G.~C. Resende, editors.
\newblock {\em Handbook of Applied Optimization}.
\newblock Oxford University Press, 2002.

\bibitem{Sutherland}
W.~A. Sutherland.
\newblock {\em Introduction to Metric and Topological Spaces}.
\newblock Oxford University Press, 1975.

\bibitem{HNG02}
C.~van Hoyweghen, B.~Naudts, and D.~E. Goldberg.
\newblock Spin-flip symmetry and synchronization.
\newblock {\em Evolutionary Computation}, 10:317--344, 2002.

\bibitem{YooKAM07}
Y.~Yoon, Y.-H. Kim, A.~Moraglio, and B.-R. Moon.
\newblock Geometric crossovers for real-code representation.
\newblock In {\em Proceedings of the Genetic and Evolutionary Computation
  Conference}, page 1539, 2007.

\end{thebibliography}
}

\end{document}